\documentclass[letterpaper, 10 pt, conference]{ieeeconf}  

\IEEEoverridecommandlockouts                              

\usepackage{epsfig} 
\usepackage{amsmath} 
\usepackage{amssymb}  
\usepackage{tabularx}
\usepackage[utf8]{inputenc}
\usepackage[T1]{fontenc}
\usepackage{textcomp}
\usepackage{gensymb}
\usepackage{multirow}
\usepackage{tikz}
\usepackage{bbding}
\usepackage{pifont}
\usepackage{wasysym}
\usepackage{amssymb}
\let\labelindent\relax

\usepackage{enumitem}
\usepackage{xcolor, soul}

\usepackage{xcolor}
\usepackage{graphicx}
\graphicspath{ {./images/}}
\usepackage{subcaption}
\usepackage{float}
\usepackage{capt-of}
\usepackage{etoolbox}
\usepackage{adjustbox}
\usepackage[colorlinks=true]{hyperref}
\usepackage{subcaption} 
\usepackage[font=small]{caption}
\usepackage{booktabs}
\usepackage{pdfpages}
\usepackage{verbatim}
\usepackage{breqn}

\usepackage{caption}
\usepackage{subcaption}
\usepackage{breqn}
\usepackage{algpseudocode}
\usepackage[linesnumbered,ruled,lined]{algorithm2e}
\usepackage{array}
\usepackage{makecell}

\usepackage{cite}
\usepackage{multirow}
\usepackage[normalem]{ulem}
\useunder{\uline}{\ul}{}

\definecolor{darkyellow}{RGB}{176, 150, 5}

\newcommand{\SelfNote}[1]{{\color{darkyellow}{(VD) \textit{\textbf{ }}: #1}}}

\title{\textbf{UAP-BEV: Uncertainty Aware Planning using Bird's Eye View Generated From Surround Monocular Images}}

\author{Vikrant Dewangan$^{1}$, Basant Sharma$^{2}$, Tushar Choudhary$^{1}$, Sarthak Sharma$^{1}$,  \\
Aakash Aanegola$^{1}$, Arun K. Singh$^{2}$, K. Madhava Krishna$^{1}$
\thanks{$^1$are with Robotics Research Center, IIIT Hyderabad, India.} 
\thanks{$^2$are with University of Tartu, Estonia}
\thanks{Source code, simulation videos and an elaborate description of our method is available at \url{https://vikr-182.github.io/UAP-BEV/}}
}

\begin{document}
\bstctlcite{Force_Etal}

\makeatletter
\let\@oldmaketitle\@maketitle
\renewcommand{\@maketitle}{\@oldmaketitle
\centering
\includegraphics[width=0.85\linewidth]{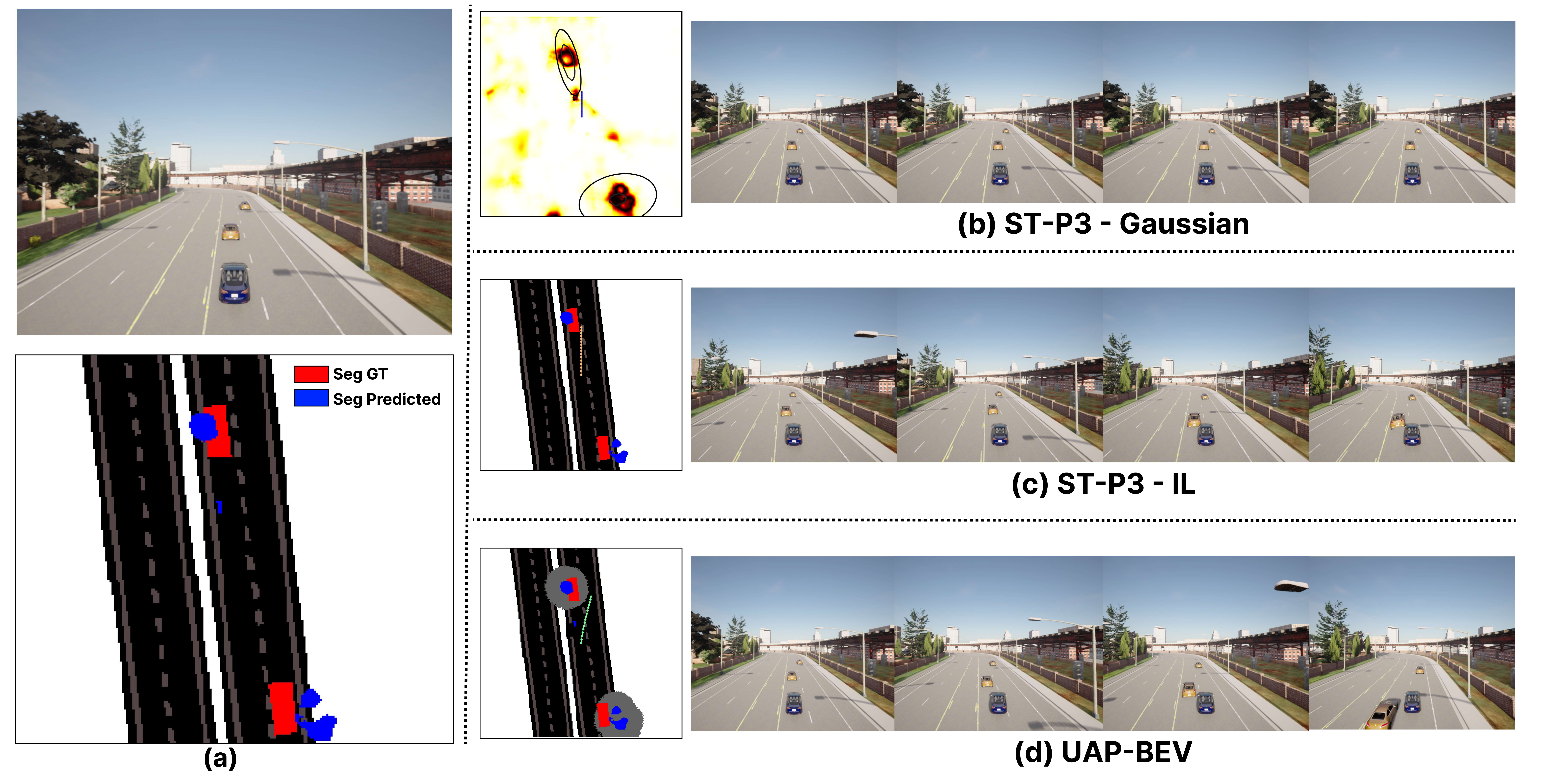}

 \captionof{figure}{\textbf{Uncertainty Aware Planner}: Given is scene simulated in CARLA \emph{\textbf{(a)}:  The ego vehicle (blue) is driving and has to overtake the static vehicle in front to reach the destination using a BEV perception model ST-P3. \textbf{(b)ST-P3 Gaussian}: Using a Gaussian approximation of underlying uncertainty proves to be conservative and the ego-vehicle fails to move ahead.  \textbf{(c) ST-P3 IL} Traditional Imitation-Learning (IL) based approaches cannot account for the error and result in a collision. Our uncertainty-aware planner \textbf{(d): UAP-BEV} is able to overtake the leading vehicle while countering the error in perception, maintaining a safe distance compared to others.} }}
\makeatother

\maketitle

\begin{abstract}
Autonomous driving requires accurate reasoning of the location of objects from raw sensor data. Recent end-to-end learning methods go from raw sensor data to a trajectory output via Bird's Eye View(BEV) segmentation as an interpretable intermediate representation. Motion planning over cost maps generated via Birds Eye View (BEV) segmentation has emerged as a prominent approach in autonomous driving. However, the current approaches have two critical gaps. First, the optimization process is simplistic and involves just evaluating a fixed set of trajectories over the cost map. The trajectory samples are not adapted based on their associated cost values. Second, the existing cost maps do not account for the uncertainty in the cost maps that can arise due to noise in RGB images, BEV annotations. As a result, these approaches can struggle in challenging scenarios where there is abrupt cut-in, stopping, overtaking, merging, etc from the neighboring vehicles. 

In this paper, we propose \textit{UAP-BEV}, a novel approach that models the noise in Spatio-Temporal BEV predictions to create an uncertainty-aware occupancy grid map. Using queries of the distance to the closest occupied cell, we obtain a sample estimate of the collision probability of the ego-vehicle. Subsequently, our approach uses gradient-free sampling-based optimization to compute low-cost trajectories over the cost map. Importantly, the sampling distribution is adapted based on the optimal cost values of the sampled trajectories. By explicitly modeling probabilistic collision avoidance in the BEV space, our approach is able to outperform the cost-map-based baselines in collision avoidance, route completion, time to completion, and smoothness. To further validate our method, we also show results on the real-world dataset NuScenes, where we report improvements in collision avoidance and smoothness. 
\end{abstract}


\section{Introduction}\label{sec:introduction}

\begin{figure*}[!htb]
\centering
\includegraphics[width=0.95\textwidth]{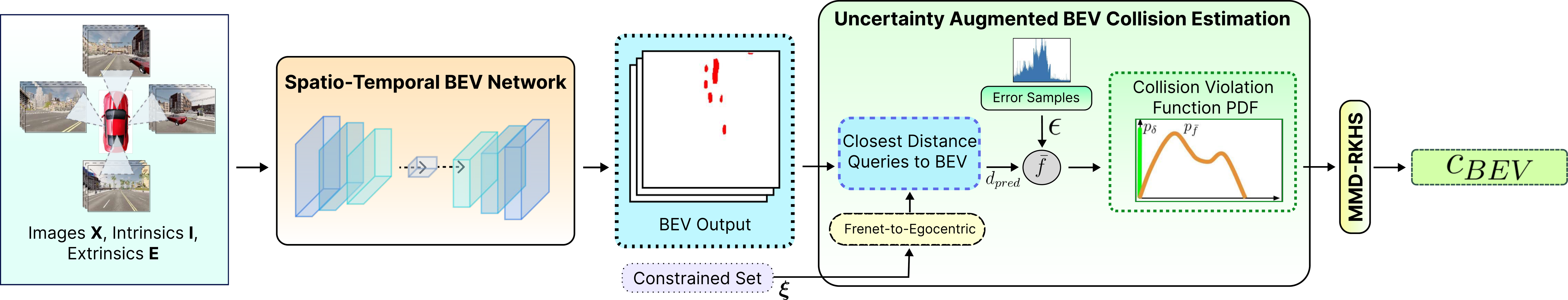}
\caption{Our approach uses a Spatio-Temporal Network \cite{stp3} to obtain a set of BEV predictions for the future which is then converted into an occupancy map prediction. We used the ground-truth information to learn the uncertainty in the occupancy map prediction. During inference time, we query the closest occupied cell to the ego vehicle and then perturb it with samples drawn from the learned uncertainty. We then use the noisy samples of distance queries and use Reproducing Kernel Hilbert Spaces (RKHS) of Probability Density Functions (PDF) of Collision Violation Function, to optimize our uncertainty-aware trajectory with the Maximum Mean discrepancy (MMD) measure as the surrogate cost for collision avoidance. We adopt a sampling-based approach and augment a projection operator into the optimization pipeline for constraint satisfaction.}
\label{fig:pipeline}
\end{figure*}

BEV representations such as Occupancy Maps or Grids have become popular in the context of Autonomous Driving \cite{lss, fiery, nmp, stp3}. The popularity of BEV stems from its appearance-agnostic characteristics, unlike direct monocular perceptual inputs. Moreover, BEV representations can effectively tap into the vast legacy of planning frameworks tailored to leverage such representations. \cite{JLB}

BEV layouts based on dense 3D LIDAR inputs are typically accurate and amenable to trajectory forecasting or layout evolution - a vital cog for motion planning in dynamic on-road scenes \cite{urtasunmp3, nmp}. The same, however, cannot be said for BEV estimation and evolution based on monocular perceptual inputs. Literature concerning the layout evolution of dynamic actors in a scene has been typically sparse \cite{stp3, fiery}. Such layouts of dynamic agents in a scene tend to be noisy and unreliable, all the more so when layout estimation and trajectory execution are interleaved in a closed-loop setting. Data-driven BEV evolution estimation typically does not consider vehicle maneuvers such as overtaking or abrupt lane changes during dataset collection. As a consequence, when such maneuvers are executed, the layout estimates become unreliable and noisy and are rendered ineffective for a motion planner relying on such estimates.

In this paper, we propose UAP-BEV, a novel Uncertainty Aware Planning in the BEV space that is competent to handle non-parametric noise inherent to data-driven BEV estimation of dynamic actors in a scene. Our algorithm in itself consists of two parts. In the first part, we sample the BEV derived from monocular cameras and obtain an uncertainty-aware estimate of the distance to the closest obstacle. This is then subsequently mapped to the probability of collision between the ego and the neighboring vehicles using the concept of distribution embedding in Reproducing Kernel Hilbert Space (RKHS). In the second part, we use a custom sampling-based optimization to compute trajectories that minimize collision avoidance probability while producing smooth trajectories. Our optimizer's novelty stems from using a projection operation that pushes the sampled trajectories toward constraint satisfaction before evaluating their costs. The constraints stem from the velocity, acceleration bounds, and barrier functions \cite{zeng2021safety} accounting for lane adherence and distance-keeping with the leading vehicles based on the BEV predictions of the neighboring vehicles.

\textbf{Contributions: }
\begin{enumerate}
    \item We propose for the first time in literature to the best of our knowledge, \textit{UAP-BEV}: an Uncertainty Aware Planner for noisy BEV layouts obtained through data-driven methods with monocular perceptual inputs.
    \item \textit{UAP-BEV} includes a novel sampling-based optimizer that can efficiently minimize the BEV-derived costs.
    \item We show significant performance enhancements in a diverse set of metrics over SOTA prior methods \cite{lss},\cite{stp3} that do not consider BEV noise 
    in simulation experiments conducted on a number of CARLA towns over a diverse set of trajectory metrics.
    \item We also show through ablations the role of barrier functions, that when utilized along with the collision probability estimates reduce collision rates to zero despite noisy BEV estimates.
\end{enumerate}

\section{Related Work} \label{relatedwork}

\noindent \textbf{End-to-End (E2E) Learning Based Approaches}: From a modular architecture comprising of a cascade of task-specific blocks (sensor fusion, perception, planning, and control), the autonomous driving stack has evolved to an E2E system \cite{neat,lss,stp3,mp3,p3,nmp} that learns to generate driving behaviours from sensor inputs like LiDARs and cameras.

Works like \cite{mp3,p3,nmp} take in voxelised LiDAR point cloud as input and use $3$D object detection \cite{nmp} or semantic BEV generation \cite{p3,mp3}  as auxiliary tasks. Learning pipelines proposed in \cite{neat,lss,stp3} take in surrounding monocular images as input to reason about the semantic BEV representation. Approaches like \cite{mp3,p3,nmp,lss,stp3} model driving behaviour by using classification loss or max-margin loss on a set of template trajectories (usually from recorded driving behaviours), where learnt imitation behaviour closest to the ground truth trajectory is encouraged. \cite{neat} predicts an offset vector towards a target waypoint which is converted to motion commands by the lower-level controllers. 

\noindent \textbf{Uncertainty In Bird's Eye View Representations}: There can be multiple sources of \textit{aleatoric uncertainty}
in BEV tasks such as the noise present in BEV annotations, intrinsic, extrinsic, and input RGB.  Approaches like \cite{kendall2017uncertainties,stp3,fiery,kendall2018multi} have aimed at quantifying uncertainties in perspective tasks like depth regression\cite{kendall2018multi,kendall2017uncertainties}, per-pixel semantic segmentation\cite{kendall2018multi,kendall2017uncertainties}, \cite{kendall2017uncertainties,kendall2018multi}. In BEV space, few
(\cite{stp3,fiery}) have addressed the issue of noise. The authors of \cite{stp3,fiery} quantify uncertainty by weighing each task's loss by its \textit{homoscedastic\ uncertainty}, taking inspiration from multi-task setting in \cite{kendall2018multi}. However, it is unclear if the \textit{aleatoric uncertainty} present in the RGB space is able to translate well into the BEV space. Further, it is not clear if \cite{kendall2018multi} is able to manage the noisy annotations in the BEV space. The authors of \cite{calib-uncertainty} fit a 2D Gaussian to every object cluster. However, this approach leads to conservatism in driving behaviour.  

\noindent \textbf{Non-parametric Uncertainty:} The above methods neither perform accurate uncertainty reasoning in the BEV, nor couple it to a planner. Modelling the error in observations as a non-parametric distribution is a popular way to deal with the uncertainty present in the inputs, and annotations and plan with it. Along this line, authors of \cite{cco-voxel} have proposed Maximum Mean Discrepancy (MMD) in RKHS as an estimate of collision probability conditioned on the sensor noise. Their gradient-free Cross Entropy Minimization (CEM) and reduced set allowed them to achieve collision avoidance, smoothness, and real-time performance improvements. \cite{urbanfly} adopts the same approach on plane segmentation tasks needed for quadrotor navigation in urban settings. Our work extends \cite{cco-voxel}, \cite{urbanfly} to dynamic and uncertain autonomous driving settings where monocular RGB images are the primary sensing modality.



%
\section{Problem Formulation}
\tiny
\begin{table*}[!t]
\centering
\caption{{List of Inequality Constraints used in the projection optimization at step $k$ }}
\begin{tabular}{|c|c|c|c|c|c|}
\hline
Constraint Type & Expression & Parameters   \\ \hline
\makecell{Discrete-time barrier\\ for longitudinal separation \\ $g_{1}$} & \makecell{$g_{2}[k]: h_{2}[k+1]-h_{2}[k]\geq -\gamma_{long} h_{2}[k]$\\ $h_{2}[k] = x_{o}[k]-x[k] \geq s_{min}$} & \makecell{$s_{min}$: minimum longitudinal separation \\ $x_{o}[k]$:x-coordinate of leading vehicle \\ at time index k\\ $\gamma_{long}$: Longitudinal barrier constant \cite{zeng2021safety} } \\ \hline
\makecell{Velocity bounds \\ $g_{2} = (g_{2, lb}, g_{2, ub} )$} & \makecell{$ g_{2, ub }[k]: \sqrt{\dot{x}[k]^2+\dot{y}[k]^2}\leq v_{max}$\\$g_{2, lb}[k]:\sqrt{\dot{x}[k]^2+\dot{y}[k]^2}\geq v_{min}$} & \makecell{$v_{min}, v_{max}$: min/max velocity\\ of the ego-vehicle}   \\ \hline
\makecell{Acceleration bounds \\$g_3$} & $g_{3}[k]: \sqrt{\ddot{x}[k]^2+\ddot{y}[k]^2}\leq a_{max}$ & \makecell{$a_{max}$: max acceleration \\of the ego-vehicle}  \\ \hline
\makecell{Discrete-time barrier \\ for Lane boundary\\ $g_{4} = (g_{4, lb}, g_{4, ub} ) $} & \makecell{$g_{4, ub}[k]: h_{4, ub}[k+1]-h_{4, ub}[k]\geq -\gamma_{lane} h_{4, ub}[k]$, \\ $g_{4, lb}[k]: h_{4, lb}[k+1]-h_{4, lb}[k]\geq -\gamma_{lane} h_{4, ub}[k]$\\ $h_{4, ub}[k] = -y[k]+y_{ub}$ \\ $h_{4, lb}[k] = y[k]-y_{lb}$} & \makecell{$y_{lb}, y_{ub}$: Lane limits as a function \\of the ego-vehicle's position.\\ $\gamma_{lane}$: Lane barrier constant \cite{zeng2021safety}} \\ \hline
\end{tabular}
\normalsize
\label{ineq_list}
\end{table*}
\normalsize
\subsubsection*{Symbols and Notations} We use lower/upper-case normal font letters to represent scalars, while bold-font small-case variants represent vectors. Matrices and tensors are represented by upper-case bold font letters.

Given a route $\mathcal{R}$ in the form of a reference centerline and the motion of other agents in the scene represented on the BEV grid, we devise a local planner to guide the ego-vehicle without collision along the centerline. Assume the availability of monocular images $\mathbf{X_{k}^n} \in \mathbb{R}^{H \times W \times 3}$ from surround $N$ camera setup of the current timestep $k_0$ and the previous $P$ steps, $ k \in \{{k_0 - P, \ldots k_0}\}$
We predict BEV representations $\mathbf{O_{k}} \in \mathbb{R}^{H \times W}$, $F$ frames into the future, $ k \in \{k_0, k_1, \ldots k_F\}$. The BEV representation is then translated into a cost map and handed to the trajectory planner. 

\subsubsection*{Frenet Frame Planning and Trajectory Parametrization} 
As is typical of autonomous driving, we perform trajectory planning in the road-aligned frame known as the Frenet frame. This frame allows us to treat curved roads as one with straight-line geometry. Additionally, the longitudinal and lateral motions of the vehicle will always be aligned with the $X$ and $Y$ axes of the Frenet frame, respectively, simplifying the optimization framework. We parameterize the $x$ and $y$ trajectories of the ego-vehicle in the Frenet frame in the following form, where $k_n$ represents the end step of the planning horizon. Note that the planning horizon need not be the same as the prediction horizon of the BEV.

\small
\begin{align}
    \begin{bmatrix}x[k_0]&\ldots&x[k_n]
    \end{bmatrix}^T = \mathbf{W}\mathbf{c}_{x},\begin{bmatrix}y[k_0]&\ldots&y[k_n]
    \end{bmatrix}^T = \mathbf{W}\mathbf{c}_{y},
    \label{param}
\end{align}
\normalsize

\noindent The matrix $\textbf{W}$ is formed by a piece-wise combination of cubic polynomial trajectories, each of which operates for a time interval $\delta t$. $\mathbf{c_x}$ and $\mathbf{c_y}$ are coefficients that are obtained through the optimization process. The higher derivatives of the position trajectory have the general form  $\mathbf{W}^{(q)}\mathbf{c}_{x}$ (along the $x$-axis), where $\mathbf{W}^{(q)}$ represents the $q^{th}$ derivative of the basis matrix. The $y$ component of the ego-vehicle trajectory and its derivatives are obtained similarly with the same basis matrices. 

We can formalize the local planner in the following form, wherein $(.)^{(q)}$ represents the $q^{th}$ derivative of the variable and $(x[k], y[k])$ represents the position of the ego-vehicle at time-step $k$.
\small{
\begin{subequations}
\begin{align}
    \sum_k c_a(x^{(q)}[k], y^{(q)}[k])+c_{BEV}(x[k], y[k]), \label{cost}\\
    (x^{(q)}[k_0], y^{(q)}[k_0], x^{(q)}[k_f], y^{(q)}[k_f]) = \textbf{b}, \label{eq_constraints}\\
    \textbf{g}(x^{(q)}[k], y^{(q)}[k]) \leq 0 \label{ineq_constraints}
\end{align}
\end{subequations}
}
\normalsize
\noindent The first term in the cost function $c_a$ captures costs that can be analytically modeled: smoothness, a departure from cruise speed, tracking, etc.
The second term $c_{BEV}$ is the cost map generated from the BEV representation. Typically, $c_{BEV}$ captures drivable area, agent interactions, etc. The equality constraints \eqref{eq_constraints} ensure boundary conditions on the position derivatives. Our formulation considers $q = (0, 1, 2)$. The inequality constraints \eqref{ineq_constraints} can capture bounds on velocities, accelerations, drivable area, etc. We present a list of inequalities contained in $\textbf{g}(.)$ in Table \ref{ineq_list}. 

\subsubsection*{Core Challenges} The core challenges associated with the BEV-based local trajectory planning can be summarized as follows

\begin{itemize}
    \item Typically, $c_{BEV}$ is obtained using large neural networks with complicated architectures \cite{stp3, lss, nmp}, making it challenging to apply gradient-based optimization for solving \eqref{cost}-\eqref{ineq_constraints}. Existing works like \cite{nmp} follow a gradient-free approach where trajectories are sampled from a distribution, and then the cost \eqref{cost} is evaluated on them. The lowest cost trajectory is chosen as the optimal one. This process represents a single iteration of a full-blown sampling-based optimizer like CEM \cite{cem} or Model Predictive Path Integral \cite{mppi}. Thus, existing works do not leverage the improvements achieved by adapting the sampling distribution online.

    \item Existing work like \cite{stp3, lss, nmp} rely purely on $c_{BEV}$ to compute optimal driving behavior. However, the existing $c_{BEV}$ is oblivious to the underlying uncertainty stemming from noisy sensors. We show later that such inadequacy reduces their performance in a closed-loop setting.  
\end{itemize}

We approximate predictions of neighboring vehicle centers that can be obtained by averaging the predicted vehicle representation in the BEV frame. This is used to formulate discrete-time barrier constraints \cite{zeng2021safety} for longitudinal separation $g_1$. 

\subsubsection*{Compact Matrix Representation} Using the trajectory parameterization\eqref{param}, we can represent \eqref{cost}-\eqref{ineq_constraints} in the following compact form. This representation will simplify the exposition in later sections.
\small{
\begin{align}
    c_a(\boldsymbol{\xi})+c_{BEV}(\boldsymbol{\xi})\\
    \textbf{A}\boldsymbol{\xi} = \textbf{b}, \qquad \textbf{g}(\boldsymbol{\xi}) \leq \textbf{0}
\end{align}
}
\normalsize
In the following two sections, we present our core algorithmic results. We first create uncertainty-aware occupancy grid maps and nominal estimates of neighboring vehicles' trajectories from BEV. Subsequently, we propose our sampling-based optimization for local trajectory planning. Note that we convert our trajectory $\boldsymbol{\xi}$ from frenet to ego-centric frame before applying the $c_{BEV}$ as the BEVs are in ego-centric frame.

\begin{figure}[!htb]
\includegraphics[width=\columnwidth]{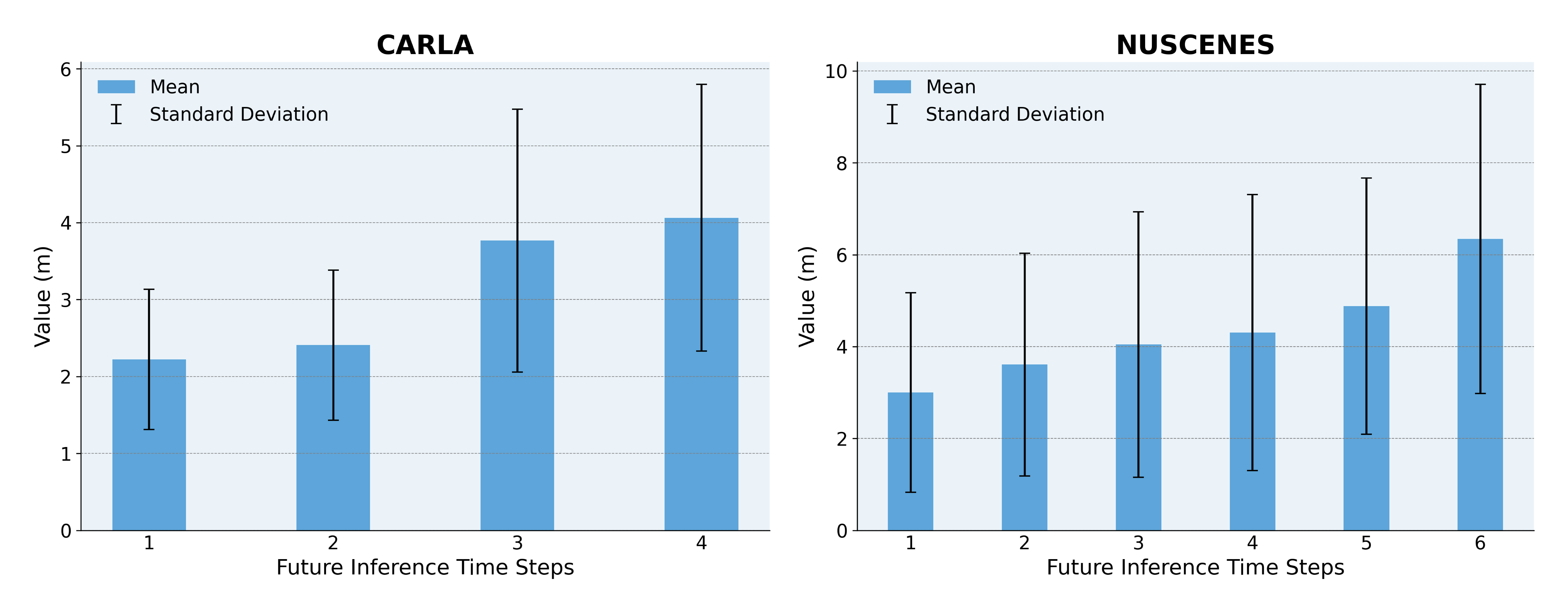}
\captionsetup{width=\columnwidth}
\caption{The mean and variance of the Error Distribution $\mathbf{\Delta_k}$ with time for CARLA and NuScenes. The mean and standard deviation increase with time, indicating that predictions further into the future are less reliable.}
\label{fig:mean-std-plot}
\end{figure}

\section{Uncertainty Aware BEV Representations}
Given a sequence of images for $N$ surround cameras, for the  $P$ past frames, $\mathbf{X}_{k}^n$,  we generate  BEV representations, for a future horizon of $F$ frames using ST-P3's architecture \cite{stp3, lss}, centered around the current ego-vehicle location. The BEV representation can be converted to an occupancy map $\textbf{O}$ from which we can obtain distance queries to the closest obstacle. More precisely, let $D_{O}: \mathbb{R}^2 \rightarrow \mathbb{R}$ be a computationally fast function such that $d[k] = D_0( (x[k], y[k]), \textbf{O}  )$ represents the distance to the closest occupied cell at any time step $k$ for any query point $(x[k], y[k])$. The distance queries over all time steps can be converted into collision costs in the form 

\small{
\begin{align}
    \overline{f} = \prod_{k = k_0}^{k = k_n} \max(r_{safe} - d[k], 0)
    \label{f_bar}
\end{align}
}
\normalsize
\noindent where $r_{safe}$ is the required minimum clearance between the ego-vehicle and its closest neighbor at time-step $k$.

We next present a core component of our pipeline; estimating uncertainty in $d[k]$ and formulating a probabilistic variant of \eqref{f_bar} that can model probabilistic safety.

\subsection{Augmenting BEV with Error on Closest-Distance Queries}
\noindent In simulators such as CARLA\cite{carla} and datasets such as NuScenes\cite{nuscenes}, we have access to the ground truth of the vehicles' positions. We can use this information to construct the so-called ground-truth BEV. Comparing the predicted occupancy map with that generated from the ground truth allows us to quantify the uncertainty in the distance queries. Fig.\ref{fig:mean-std-plot} shows the mean and standard deviation of the error in distance to the closest occupied cell $d[k]$ observed in CARLA and NuSenes. As can be seen, due to the nature of BEV segmentation provided by  ST-P3's architecture, the uncertainty in distance queries increases with time. We fit an average time-dependent distribution $\boldsymbol{\Delta}_k$ over several scenes in CARLA and NuScenes.

\subsubsection{Probabilistic Safety Through Distance Samples}
\noindent Let $\epsilon_{k, i}$ be the $i^{th}$ sample drawn from $\boldsymbol{\Delta}_k$ such that $d_{i}[k] = d[k]+\epsilon_{k, i}$ represent the noisy samples of the distance to the closest occupied cell. We draw $m$ such samples. We can now use these distance samples to compute the sample estimate of collision-cost \eqref{f_bar} in the following manner :

\small{
\begin{align}
    \overline{f}_i = \prod_{k = k_0}^{k = k_n} \max(r_{safe} - d_i[k], 0)
    \label{f_bar_samples}
\end{align}
}
\normalsize

\noindent Expression \eqref{f_bar_samples} represents the various possibilities of collision cost due to the uncertain distance information. We intend to use all the samples of $\overline{f}_i$ to infer some notion of probability of collision avoidance. A simple choice is to just compute the mean of all the samples. However, such an approach would not capture the true notion of risk. In our approach, we use the concept of distribution embedding in RKHS. 

Let $\kappa: \mathbb{R}^2 \rightarrow \mathbb{R}$ be a positive-definite kernel function (e.g. Gaussian kernel) associated with RKHS. Then, the RKHS embedding of $\overline{f}_i$ is given by \cite{cco-voxel}.

\small{
\begin{align}
    \mu_{\overline{f}} &= \sum_{i=0}^{m} \frac{1}{m} \kappa(\bar{f}_i, \cdot), \qquad \mu_{{\delta}} = 
 \sum_{i=0}^{m} \frac{1}{m} \kappa(\mathbf{0}, \cdot)     
    \label{mu_f_bar}
\end{align}
}
\normalsize

\noindent The first half of  \eqref{mu_f_bar} use all the samples of $f_i$ to represent the underlying distribution as a point $\mu_{\overline{f}}$ in RKHS. As shown in \cite{cco-voxel}, \cite{urbanfly}, the ${l}_2$ distance between $\mu_{\overline{f}}$ and the RKHS embedding of Dirac-Delta distribution centered at zero, $\mu_{{\delta}}$ can be used as a measure of the probability of collision avoidance. Thus, we define our uncertainty-aware $c_{BEV}$ as 
\small{
\begin{align}
    c_{BEV} =  \overbrace{\Vert\mu_{\overline{f}} - \mu_{{\delta}}\Vert_2^2}^{MMD}
    \label{uncert_bev}
\end{align}
}
\normalsize
Few important points about $c_{BEV}$ are in order

\begin{itemize}
    \item First, the r.h.s of \eqref{uncert_bev} can be efficiently computed using the so-called kernel trick. Also, \eqref{uncert_bev} explicitly depends on the ego-vehicle trajectory. 
    \item Given a set of ego-vehicle trajectories, the one for which the $c_{BEV}$ of \eqref{uncert_bev} is close to zero, will have the highest probability of collision avoidance.
\end{itemize} 

In the next section, we present our sampling-based optimizer to compute trajectories optimal with respect to our uncertainty-aware $c_{BEV}$

\section{Sampling-Based Optimization for Local Planning}
\label{sec:path-planning}
The developments in the last section provide us $c_{BEV}$ as an MMD map computed through a neural network-based BEV representation augmented with uncertainty estimates. This section develops a sampling-based approach for optimizing over $c_{BEV}$. The key novelty of our optimizer is that it incorporates a projection operator to push the sampled trajectories toward feasible regions before evaluating cost over them. 

\subsection{Proposed Optimizer}

\noindent The overall algorithm is presented in Alg.\ref{algo_1}, wherein the left superscript $l$ is used to track the values of the respective variable across iterations. The algorithm proceeds by sampling behavioral inputs $\textbf{p}_j$ such as lateral offsets and desired velocity setpoints. These are then fed to a Frenet space planner inspired by \cite{wei2014behavioral}. The trajectory coefficients that the Frenet planner returns are then fed to our projection optimizer in lines 7-8. The resulting output $\overline{\boldsymbol{\xi}_j}$ is then evaluated for constraint residuals in line 9. We rank the top $n_{s}$ samples with the lowest constraint residual in the $ConstraintEliteSet$ in line 10. In line 11, we compute an augmented cost (residuals+primary cost) over the samples from $ConstraintEliteSet$. We then rank these samples based on the augmented cost value, extract the lowest $n_e$ samples, and place them in $EliteSet$ (line 13). We update the sampling distribution based on the samples from the $EliteSet$. Specifically, we use the formula \eqref{mean_update}-\eqref{cov_update} from \cite{bhardwaj2022storm} to update the mean and covariance for sampling in the next iteration. The constant $\beta$ is the so-called temperature parameter.

\small{
\begin{subequations}
\begin{align}
    {^{l+1}}\boldsymbol{\mu}_{\textbf{p}} = (1-\eta){^{l}}\boldsymbol{\mu}_{\textbf{p}}+\eta\frac{\sum_{j=1}^{j=n_{e}} s_j\textbf{p}_j   }{\sum_{j=1}^{j=n_{e}} s_j}, \label{mean_update}\\
    {^{l+1}}\boldsymbol{\Sigma}_{\textbf{p}} = (1-\eta){^{l}}\boldsymbol{\Sigma}_{\textbf{p}}+\eta\frac{ \sum_{j=1}^{j=n_{e}} s_j(\textbf{d}_j-{^{l+1}}\boldsymbol{\mu}_{\textbf{p}})(\textbf{p}_j-{^{l+1}}\boldsymbol{\mu}_{\textbf{p}})^T}   {\sum_{j=1}^{j=n_e} s_j} \label{cov_update}\\
    s_j = \exp{\frac{-1}{\beta}(c_{a}(\overline{\boldsymbol{\xi}}_j )+c_{BEV}(\overline{\boldsymbol{\xi}}_j )+r_j(\overline{\boldsymbol{\xi}}_j }) \label{s_formula}
\end{align}
\end{subequations}
}
\small{
\noindent 
\begin{algorithm}[!h]
\caption{Sampling-Based Optimization Over Uncertainty Augmented Learned BEV-based Costs}
\small
\label{algo_1}
\SetAlgoLined
$N$ = \textbf{Maximum number of iterations}\\
Initiate mean $^{l}\boldsymbol{\mu}_{\textbf{p}}, ^{l}\boldsymbol{\Sigma}_{\textbf{p}}$, at $l=0$ for sampling Frenet Parameters $\textbf{p}$\\
\For{$l=1, l \leq N, l++$}
{
Draw $\overline{n}_{s}$ Samples $(\textbf{p}_1, \textbf{p}_2, \textbf{p}_3, ...., \textbf{p}_{n_s})$ from $\mathcal{N}(^{l}\boldsymbol{\mu}_{\textbf{p}}, ^{l}\boldsymbol{\Sigma}_{\textbf{p}})$\\
Initialize $CostList$ = []\\

Query Frenet-planner for $\forall \textbf{p}_j$: $\boldsymbol{\xi}_j = \text{FrenetPlanner}(\textbf{p}_j)$ \\
\text{Project to Constrained Set} \\
\small
\begin{align*}
    \overline{\boldsymbol{\xi}}_j = \arg\min_{\overline{\boldsymbol{\xi}}_j} \frac{1}{2} \left\Vert \overline{\boldsymbol{\xi}}_j-\boldsymbol{\xi}_j \right \Vert_2^2\\
     \textbf{A}\overline{\boldsymbol{\xi}}_j = \textbf{b}, \qquad \textbf{g}(\overline{\boldsymbol{\xi}}_j) \leq \textbf{0}
\end{align*}
\normalsize

Define constraint residuals: $r_j(\overline{\boldsymbol{\xi}}_j)$\\
$ConstraintEliteSet  \gets$ Select top $n_{s}$   samples of $\textbf{p}_j, \overline{\boldsymbol{\xi}}_j$ with lowest constraint residual norm.\\
$cost \gets$ $c_{a}(\overline{\boldsymbol{\xi}}_j)+c_{BEV}(\overline{\boldsymbol{\xi}}_j)+r_j(\overline{\boldsymbol{\xi}}_j)$,  over $ConstraintEliteSet$ \\
append ${cost}$ to $CostList$ \\
          
$EliteSet  \gets$ Select top $n_{e}$ samples of ($ \textbf{p}_j, \overline{\boldsymbol{\xi}\textbf{}}_j $)  with lowest cost from $CostList$.\\

$({^{l+1}}\boldsymbol{\mu}_{\textbf{p}}, {^{l+1}}\boldsymbol{\Sigma}_{\textbf{p}} ) \gets$ Update distribution based on $EliteSet$
}
\normalsize
\Return{ Trajectory coefficients $\overline{\boldsymbol{\xi}}_j$ corresponding to lowest cost in the $EliteSet$}
\normalsize
\end{algorithm}
}
\normalsize

\subsection{Batch Projection}

\noindent The optimization in lines 7-8 can be done in parallel. We extended the GPU accelerated projection optimizer from \cite{masnavi2022visibility}, \cite{singh2022bi} to include the discrete-time barrier constraints for longitudinal separation and lane-boundary constraints (recall Table \ref{ineq_list}, rows 1 and 4). Moreover, unlike \cite{singh2022bi}, our projection operator does not have collision constraints as these have been rolled into the $c_{BEV}$ costs. We present the detailed derivation in the appendix. But the core idea essentially boils downs to reducing projection in lines 7-8 to a sequence of optimizations that can all be trivially batched over GPUs.

\section{Experiments}
\label{sec:results}
\begin{figure*}
\centering
\includegraphics[width=0.85\textwidth]{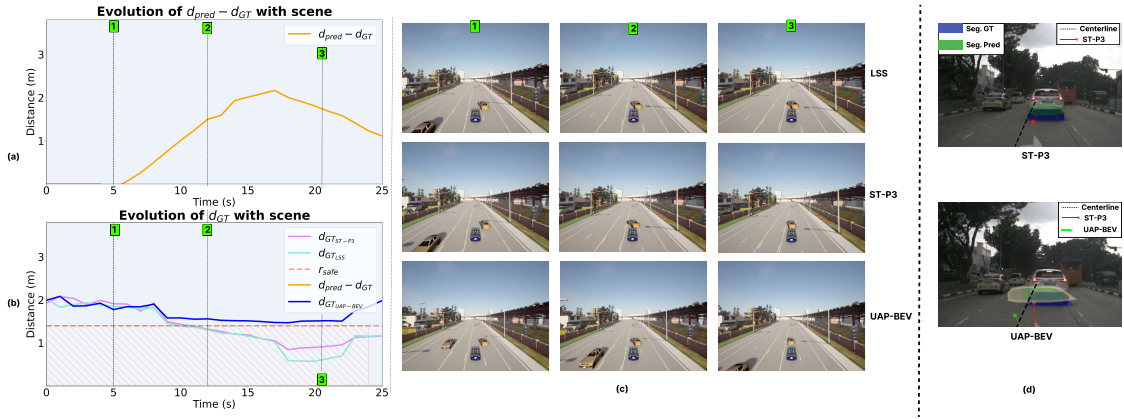}
\caption{Given is a cutin scenario generated through the ST-P3 pipeline. Fig. (a) represents the error in perception that occurs for the leading vehicle during the cut-in. In Fig.(b) it can be seen there is a difference between the ground truth ($d_{GT}$) and the predicted ($d_{pred}$) distance to the closest occupied cell at different time instants. As a result, the ST-P3 trajectory leads to a collision, denoted by the $d_{GT}$ going below $r_{safe}$. In contrast, our approach successfully navigates the cutin. This result is visualized through scene images in Fig. (c). Fig (d) visualizes a scene in NuScenes where the effect of uncertainty augmentation (yellow) can be seen.} \label{fig:qualitative-1}
\end{figure*}

\subsection{Implementation Details}

We conduct our closed-loop experiments on the self-driving simulator CARLA \cite{dosovitskiy2017carla} v0.9.13, owing to its high-precision physics engine and sensors capable to our use case. We implemented our sampling-based optimizer in Google's JAX - GPU accelerated internal library\cite{jax2018github}.  Our perception models were trained on Nvidia 3090, and the CARLA simulation and open-loop experiments on NuScenes were conducted on Nvidia TITAN X.  We used $\gamma=0.1$ for the RBF Kernel. We used $\gamma_{lane}=0.9$ and $\gamma_{long}=0.9$.  The temperature parameter $\beta$ was taken as 0.9. and the learning-rate $\eta$ at $0.6$. We chose $v_{max}, v_{min}$ at $(10, 0)$.
We adopted the BEV configuration setting of ST-P3, with $(200, 200)$ grid at 0.20m $\times$ 0.20m resolution in CARLA, and $0.50m \times 0.50m$ in NuScenes. We use $P = 3$ time frames of past context and predict $F=4$ frames into the future for CARLA and $F=6$ frames for NuScenes.  
\subsection{CARLA Scenarios Setup}
We divide our experiments on CARLA into 2 categories based on the lane boundary  - Inlane Driving (\textbf{In}) and Overtaking Allowed \textbf{Ov}. Within Inlane, we create Abrupt Stopping/Slowing and Cutin. We change $y_{lb}=3.5$  to allow overtaking. \noindent We adopt the 14 standard routes introduced in \cite{neat} and test on 2.8 km of route length across multiple towns. Due to the controlled nature of the neighboring vehicles, we can re-create the same scenario for the baselines discussed in the next subsection. 

\subsection{Baselines and Metrics}
We use camera-based methods in our approach which leverage BEV representations as intermediates for planning. Note that we do not consider \cite{cco-voxel, urbanfly} as they parameterize the trajectories over 3D coefficients, making them unsuitable for AD applications. We also do not consider \textbf{ST-P3 Gaussian} \cite{calib-uncertainty} as we found it to be too conservative and unable to complete the route.
\begin{enumerate}[label=(\alph*)]
    \item Lift-Splat-Shoot (Static) (LSS) \cite{lss}: Lift-Splat Shoot used to generate BEV representations of the current timestep, with a cost grid $c_{LSS} \in \mathbb{R}^{H \times W} = (200, 200)$. At test time, we sample polynomial trajectories from a fixed set and the lowest cost trajectory is passed to the PID controller to obtain the control inputs.
   \item ST-P3 Imitation Learning (IL) (ST-P3-IL) \cite{stp3}: BEV representations of the current and future timesteps are generated. It predicts a Cost-Volume over time, $c_{ST-P3} \in \mathbb{R}^{F\times H \times W}=(4, 200, 200)$. At test time, we use the same approach as stated above.  
\item ST-P3 with Sampling-based optimization (ST-P3-SO): Here, we use our sampling-based optimization Alg. \ref{algo_1} over the cost volume to obtain the optimal trajectory. The $c_{BEV}$ cost is the cost-volume generated by the ST-P3. Thus, this baseline has been constructed to showcase the importance of our proposed uncertainty-aware $c_{BEV}$ based on MMD.
\end{enumerate}


The metrics in closed-loop CARLA simulation are -
\begin{enumerate}
    \item \textbf{Collisions}: The average no. of vehicle collisions per km.
    \item \textbf{Route Completion (RC)}: The percentage of routes completed, i.e: the ego-vehicle navigates to the target, without getting stuck or stopping for a fixed time.
    \item \textbf{Duration}: Total time (in s) for successful route.
    \item \textbf{Smoothness}: Rate of Change of Acceleration $(m/s^3)$.
\end{enumerate}

For NuScenes, we perform open-loop experiments and report Collision Rate (\%) and Smoothness.
\subsection{Qualitative Results}
\noindent Fig.\ref{fig:qualitative-1} shows a qualitative comparison between ST-P3 and our approach. In this scenario, the ego vehicle experiences a cutin from a vehicle in the adjacent lane. Due to a mismatch between the ground-truth $d_{GT}$ and predicted $d_{pred}$ distance to the closest obstacle, the ST-P3 trajectory collides with the leading vehicle. In contrast, our uncertainty-aware approach (UAP-BEV) that models $c_{BEV}$ through MMD and uses Alg.\ref{algo_1} for optimization, is able to successfully navigate through the cutin. A result on NuScenes is also visualized. In the next subsection, we further cement the observations made here with our quantitative benchmarking.

\subsection{Quantitative Results}

\begin{table}[!htb]
\centering
\scriptsize
\begin{tabular}{@{}lllrlrlrl@{}}
\toprule
\multicolumn{1}{c}{\multirow{3}{*}{\textbf{Method}}} & \multicolumn{2}{c}{\textbf{Collisions  $\downarrow$}}                     & \multicolumn{2}{c}{\textbf{RC}  $\uparrow$}                             & \multicolumn{2}{c}{\textbf{Duration  $\downarrow$}}                       & \multicolumn{2}{c}{\textbf{Smoothness  $\downarrow$}}                     \\ 
\multicolumn{1}{c}{}                        & \multicolumn{2}{c}{(/km)}                          & \multicolumn{2}{c}{(\%)}                        & \multicolumn{2}{c}{(s)}                            & \multicolumn{2}{c}{(jerk,$m/s^3$)}                        \\ \cmidrule(l){2-9} 
\multicolumn{1}{c}{}                 & \multicolumn{1}{c}{\textit{In}} & \multicolumn{1}{c}{\textit{Ov}} & \multicolumn{1}{c}{\textit{In}} & \multicolumn{1}{c}{\textit{Ov}} & \multicolumn{1}{c}{\textit{In}} & \multicolumn{1}{c}{\textit{Ov}} & \multicolumn{1}{c}{\textit{In}} & \multicolumn{1}{c}{\textit{Ov}} \\ \midrule
LSS                           & 0.013                    & 0.014                    & 26.57                    & 20                       & 39.84                    & 31.84                    & 4.74                     & 4.85                     \\ 
ST-P3                                  & 0.011                    & 0.013                    & 48.44                    & 46                       & 28.56                    & 26.34                    & 4.53                     & 4.55                     \\ 
ST-P3-SO                              & 0.004                    & 0.005                    & 85.94                    & 90                       & 22.19                    & 22.7                     & 2.98                     & 2.95                     \\ 
UAP-BEV                                  & $\sim$ 0                        & $\sim$ 0                        & \textbf{100}             & 100                      & \textbf{18.9}            & 19.1                     & \textbf{1.88}            & 2.93                     \\ \cmidrule(l){1-9} 
\end{tabular}
\caption{Benchmark Comparison on CARLA on Inlane (\textit{In}) and Overtaking Scenarios (\textit{Ov}).}
\label{table:benchmark}
\end{table}
\normalsize



\noindent Table \ref{table:benchmark} presents the quantitative results on the CARLA simulator. It can be seen that the ST-P3-SO baseline, which couples ST-P3 perception with our sampling-based optimizer from Alg.\ref{algo_1}, already substantially improves the collision rate. Moreover, our primary method UAP-BEV which uses the optimizer Alg.\ref{algo_1} with our MMD based $c_{BEV}$ drives the collision rate to zero. Our UAP-BEV achieves an improvement of \textbf{39.8\%} in smoothness metric over all the baselines. In both in-lane driving and overtaking scenarios, our approaches achieve a route competition rate of almost double the other baselines. The ST-P3-SO and UAP-BEV demonstrate an improvement of \textbf{1.774x} and \textbf{2.06x} respectively over ST-P3-IL in route competition metric.

\begin{table}[!htb]
\centering
\begin{tabular}{c c c c c c c c}
\hline
{\textbf{Method}} & {\textbf{Collision Rate(\%) $\downarrow$}} & \textbf{Smoothness} (jerk,\textbf{$m/s^3$}) $\downarrow$  \\
\hline
LSS & 9.31 & 5.31\\ 
ST-P3 & 6.01 & 5.11\\
ST-P3-SO & 4.31 & 2.82\\
UAV-BEP & \textbf{3.34} & \textbf{2.80} \\ \hline
\end{tabular}
\caption{Benchmark Comparison on NuScenes}
\label{tab:benchmark-nuscenes}
\end{table}
\normalsize

Table \ref{tab:benchmark-nuscenes} presents results on the NuScenes dataset. Here again, our UAP-BEV shows almost two to three times reduction in a collision over LSS and ST-P3-IL. Similar improvements can be found in the smoothness metric as well. We recall that NuScenes dataset only allows us to perform open-loop simulation (executed trajectory of the ego-vehicle is pre-decided and fixed). Thus the improvement here is less drastic compared to CARLA. We also do not report the route-competition metric as it is  irrelevant here. 



\subsection{Longitudinal Barrier Ablations}

\noindent Table \ref{table:barrier} demonstrates the effectiveness of the longitudinal barrier constraints (recall Table \ref{ineq_list}). This constraint ensures a minimum separation distance from the leading vehicle in the scene. Note that we only utilize this constraint during the Inlane driving scenarios. The trajectory of the leading vehicle was obtained by computing the approximate centers of the BEV predictions. 

\begin{table}[!htb]
\centering
\scriptsize
\begin{tabular}{@{}llllll@{}}
\toprule
ID & UAP & Barrier & Collisions /km & Route Completion & Duration (s) \\ \midrule
1  &  &      & 0.004          & 85.84            & 22.43        \\
2  &  & \checkmark      & 0.0005         & 91.23            & 25.43        \\
3  & \checkmark  &      & 0.0003         & 96.37            & \textbf{19.17}        \\
4  & \checkmark  & \checkmark     & \textbf{0.0}            & \textbf{100}              & 21.18        \\ \bottomrule
\end{tabular}
\caption{With and Without Barrier}
\label{table:barrier}
\end{table}
The longitudinal barrier constraint encourages conservative driving, by rewarding trajectories that maintain a minimum distance from the leading vehicle. It leads to safer trajectories observed through reduced collisions and increased RC. However, this comes at the expense of higher execution times.

\section{Conclusion}
\label{sec:conclusions}

UAP-BEV, to the best of our knowledge, is the first method to characterize non-parametric uncertainty in the BEV frame and leverage it to compute collision-free trajectories. The proposed planner utilizes a sampling-based optimization with a novel uncertainty-aware collision cost constructed from BEV predictions. Our proposed optimizer also includes a projection operator to push the sampled trajectories towards feasible regions before computing the cost over them.
The efficacy of UAP-BEV is demonstrated through the significant performance gain over prior SoTA work \cite{lss, stp3} in closed-loop simulation on CARLA on challenging scenarios. 

\section{Appendix}
In this section, we provide further details on our projection optimizer. We begin by re-writing the velocity and acceleration bounds from Table \ref{ineq_list} in the following form.

\small
\begin{align}
    \textbf{f}_{v} = \left \{ \begin{array}{lcr}
\dot{x}[k] -d_{v}[k]\cos\alpha_{v}[k] \\
\dot{y}[k] -d_{v}[k]\sin\alpha_{v}[k]\\ 
\end{array} \right \}, v_{min}\leq d_{v}[k]\leq v_{max}
\label{vel_bound_proposed}
\end{align}
\normalsize

\vspace{-0.25cm}
\small
\begin{align}
    \textbf{f}_{a} = \left \{ \begin{array}{lcr}
\ddot{x}[k] -d_{a}[k]\cos\alpha_{a}[k] \\
\ddot{y}[k] -d_{a}[k]\sin\alpha_{a}[k]\\ 
\end{array} \right \}, 0\leq d_{a}[k]\leq a_{max}
\label{acc_bound_proposed}
\end{align}
\normalsize
The variables $\alpha_v[k], \alpha_a[k], d_v[k], d_a[k] $ will be obtained by the projection optimizer along with $\overline{\boldsymbol{\xi}}$.
\subsection{Reformulated Projection Optimization}

\noindent Using the developments in the previous section and the trajectory parametrization presented in \eqref{param}, we can now replace the projection optimization \eqref{cost}-\eqref{ineq_constraints} with the following. Note that \eqref{barrier_reform} is the matrix representation of the barrier constraints presented in Table \ref{ineq_list}.

\small
\begin{subequations}
\begin{align}
    \overline{\boldsymbol{\xi}}_j^{*} = \arg\min_{\overline{\boldsymbol{\xi}}^*_j}\frac{1}{2}\Vert \overline{\boldsymbol{\xi}}^*_j-{\boldsymbol{\xi}}_j^*\Vert_2^2\label{cost_reform}  \\
    \textbf{A} \overline{\boldsymbol{\xi}}^*_j= \textbf{b}(\textbf{p}_j) \label{eq_reform} \\
    \widetilde{\textbf{F}} \hspace{0.05cm} \overline{\boldsymbol{\xi}}^*_j = \widetilde{\textbf{e}}(\boldsymbol{\alpha}_j, \textbf{d}_j) \label{nonconvex_reform}  \\
    \textbf{d}_{min} \leq \textbf{d}_j\leq \textbf{d}_{max} \label{d_reform_1}\\
     \textbf{G}\overline{\boldsymbol{\xi}}^*_j \leq \textbf{b}_{barrier} \label{barrier_reform}
\end{align}
\end{subequations}
\normalsize
\vspace{-0.4cm}
\small
\begin{align*}
    \widetilde{\textbf{F}} &= \begin{bmatrix}
    \begin{bmatrix}
    \dot{\textbf{W}}\\
    \ddot{\textbf{W}}
    \end{bmatrix} & \textbf{0}\\
    \textbf{0} & \begin{bmatrix}
    \dot{\textbf{W}}\\
    \ddot{\textbf{W}}
    \end{bmatrix} 
    \end{bmatrix}, \widetilde{\textbf{e}} = \begin{bmatrix}
     \textbf{d}_{v, j}\cos\boldsymbol{\alpha}_{v, j}\\
  \textbf{d}_{a, j}\cos\boldsymbol{\alpha}_{a, j}\\
     \textbf{d}_{v, j}\sin\boldsymbol{\alpha}_{v, j}\\
  \textbf{d}_{a, j}\sin\boldsymbol{\alpha}_{a, j}
    \end{bmatrix} \\
    \boldsymbol{\alpha}_j &= (\boldsymbol{\alpha}_{a,j}, \boldsymbol{\alpha}_{v,j}), \mathbf{d}_j =  (\mathbf{d}_{v, j}, \mathbf{d}_{a, j}) \\
    \textbf{G} &= \begin{bmatrix}
        \textbf{G}_{ub} & \textbf{G}_{lb} & \textbf{G}_{long} \end{bmatrix}^T
, \textbf{b}_{barrier} = \begin{bmatrix}
        \textbf{b}_{ub} & \textbf{b}_{lb} & \textbf{b}_{long}
    \end{bmatrix}^T \\
    \mathbf{G}_{ub} &= \mathbf{W}_{1, ub} + (\gamma_{lane} -1)\mathbf{W}_{0, ub}, \mathbf{b}_{ub} = \gamma_{lane} y_{ub} \mathbf{1}_{(n-1) \times 1} \\
    \mathbf{G}_{lb} &= \mathbf{W}_{1, lb} + (1 - \gamma_{lane})\mathbf{W}_{0, lb}, \mathbf{b}_{lb} = -\gamma_{lane} y_{lb} \mathbf{1}_{(n-1) \times 1} \\
    \mathbf{G}_{long} &= \mathbf{W}_{1, long} + (\gamma_{long} -1)\mathbf{W}_{0, long} \\
    \mathbf{b}_{long} &= \mathbf{x}_{o, [1:n]} + (\gamma_{long}-1)\mathbf{x}_{o, [0:n-1]} - \gamma_{long}s_{min}\mathbf{1}_{(n-1) \times 1} \\
    \mathbf{W}_{1, ub} &= \mathbf{W}_{1, long} = \mathbf{W}_{[1:n]}, \mathbf{W}_{1, lb} = -\mathbf{W}_{[1:n]} \\
    \mathbf{W}_{0, ub} &= \mathbf{W}_{0, lb} = \mathbf{W}_{0, long}  = \mathbf{W}_{[0:n-1]}
\end{align*}
\normalsize
Here $\mathbf{W}_{[i:j]}$ and $\mathbf{x}_{o, [i:j]}$ represent rows $i$ to $j$ (both included) of $\mathbf{W}$ and $\mathbf{x}_{o}$,respectively.Note that $\mathbf{x}_{o}$ is obtained after stacking $x_{o}[k]$(as in Table \ref{ineq_list}) for all time instants $k=1,2,...n$.Constraints \eqref{nonconvex_reform}-\eqref{barrier_reform} act as substitutes for $\textbf{g}(\overline{\boldsymbol{\xi}}_j^*)\leq 0 $ in the projection optimization \eqref{cost}-\eqref{ineq_constraints}. 

The vectors $\boldsymbol{\alpha}_{v}, \boldsymbol{\alpha}_{a}, \boldsymbol{d}_{v}, \boldsymbol{d}_{a}$ are formed by appropriately stacking $\alpha_{v}[k], \alpha_{a}[k], d_{v}[k], d_{a}[k]$ at different time instants.Vectors $\boldsymbol{d}_{min}, \boldsymbol{d}_{max}$ are formed by stacking the lower and upper bounds for $d_a[k], d_v[k]$.
\vspace{-0.2cm}
\subsection{Solution Process} We relax the non-convex equality \eqref{nonconvex_reform} and affine inequality constraints as $l_2$ penalties and augment them into the projection cost \eqref{cost_reform}.
\vspace{-0.2cm}
\small
\begin{align} \label{aug_lag}
    \mathcal{L} &= \frac{1}{2}\left\Vert \overline{\boldsymbol{\xi}}^*_j-\boldsymbol{\xi}^*_j\right\Vert_2^2- \boldsymbol{\lambda}_{j}^T \overline{\boldsymbol{\xi}}^*_j+\frac{\rho}{2} \left \Vert \widetilde{\textbf{F}} \overline{\boldsymbol{\xi}}^*_j-\widetilde{\textbf{e}}\right \Vert_2^2\\
    &+ \frac{\rho}{2}\left \Vert \mathbf{G} \overline{\boldsymbol{\xi}}^*_{j} - \textbf{b}_{barrier} + \mathbf{s}_j \right \Vert^2 \nonumber \\ &= \frac{1}{2}\left\Vert \overline{\boldsymbol{\xi}}^*_j-\boldsymbol{\xi}^*_j\right\Vert_2^2-\boldsymbol{\lambda}_{j}^T \overline{\boldsymbol{\xi}}^*_j+\frac{\rho}{2} \left \Vert \textbf{F} \overline{\boldsymbol{\xi}}^*_j-\textbf{e}\right \Vert_2^2 \nonumber
     \\
    \textbf{F} &= \begin{bmatrix}
        \widetilde{\textbf{F}}\\
        \textbf{G}
    \end{bmatrix}, \textbf{e} = \begin{bmatrix}
        \widetilde{\textbf{e}}\\
        \textbf{b}_{barrier}-\textbf{s}_j
    \end{bmatrix}
\end{align}
\normalsize
\noindent Note, the introduction of the Lagrange multiplier $\boldsymbol{\lambda}$ that drives the residual of the second and third quadratic penalties to zero. We minimize \eqref{aug_lag} subject to \eqref{eq_reform} through Alternating Minimization (AM), which reduces to the following steps \cite{masnavi2022visibility}.

\vspace{-0.5cm}
\small
\begin{subequations}
    \begin{align}
        {^{u+1}\boldsymbol{\alpha}_j} = \arg\min_{\boldsymbol{\alpha}_j} \mathcal{L}({^u}\overline{\boldsymbol{\xi}}_j^*, {^u}\textbf{d}_j, \boldsymbol{\alpha}_j {^u}\boldsymbol{\lambda}_j, {^u}\textbf{s}_j ) \label{am_alpha}\\
        {^{u+1}\textbf{d}_j} = \arg\min_{\textbf{d}_j} \mathcal{L}({^u}\overline{\boldsymbol{\xi}}_j^*, \textbf{d}_j, {^{u+1}}\boldsymbol{\alpha}_j, {^u}\boldsymbol{\lambda}_j, {^u}\textbf{s}_j) \label{am_d} \\ 
        {^{u+1}}\mathbf{s}_{j} =\text{max}\left(0, -\mathbf{G} {^{u}}\overline{\boldsymbol{\xi}}_{j}^* - \textbf{b}_{barrier}\right) \label{am_s} \\
        {^{u+1}}\boldsymbol{\lambda}_j = {^{u}}\boldsymbol{\lambda}_j+\rho\textbf{F}^T (\textbf{F}\hspace{0.05cm} {^u}\overline{\boldsymbol{\xi}}_j^*-{^{u}}\textbf{e}_j  ) \label{am_lambda} \\
        {^{u+1}}\textbf{e}_j = \begin{bmatrix}
        \widetilde{\textbf{e}} ({^{u+1}} \boldsymbol{\alpha}_j, {^{u+1}}\textbf{d}_j ) \label{am_e} \\
        \textbf{b}_{barrier}-{^{u+1}}\textbf{s}_j
    \end{bmatrix}\\
        {^{u+1}}\overline{\boldsymbol{\xi}}_j^* = \arg\min_{\overline{\boldsymbol{\xi}}_j^*}\mathcal{L}(\overline{\boldsymbol{\xi}}_j^*, {^{u+1}}\boldsymbol{\lambda}_j, {^{u+1}}\textbf{e}_j ) \label{am_xi}
    \end{align}
\end{subequations}
\normalsize
Here $u$ is the iteration index of the projection algorithm. Optimizations \eqref{am_alpha}-\eqref{am_d} have closed-form solutions that can be batched over GPUs. The same holds from \eqref{am_s}-\eqref{am_e}. Step \eqref{am_xi} is an equality constrained QP with a batchable structure similar to that presented in \cite{masnavi2022visibility}.
\noindent



\bibliography{references}
\bibliographystyle{IEEEtran}

\end{document}